\pdfoutput=1

\documentclass[11pt]{article}

\usepackage[preprint]{acl}

\usepackage{times}
\usepackage{latexsym}

\usepackage[T1]{fontenc}

\usepackage[utf8]{inputenc}

\usepackage{microtype}
\usepackage{xspace}
\usepackage{booktabs}

\usepackage{inconsolata}

\usepackage{graphicx}

\usepackage{color, colortbl}
\definecolor{LightCyan}{rgb}{0.88,1,1}
\usepackage{multirow}

\newcommand{\ours}{Dynamic Block-Sparse Attention\xspace}

\newcommand{\reticl}{Retrieval ICL\xspace}

\newcommand{\fixedicl}{Fixed ICL\xspace}

\newcommand{\ft}{finetuning\xspace}

\newcommand{\abbrours}{DBSA\xspace}
\title{Efficient Many-Shot In-Context Learning \\ with Dynamic Block-Sparse Attention}

\author{
    \textbf{Emily Xiao}, \textbf{Chin-Jou Li},  \textbf{Yilin Zhang}, \textbf{Graham Neubig}, \textbf{Amanda Bertsch} \\
    Language Technologies Institute, Carnegie Mellon University \\
  \texttt{\{emilyx,chinjoul,jasonzh3,gneubig,abertsch\}@andrew.cmu.edu}
}

\begin{document}
\maketitle
\begin{abstract}
Many-shot in-context learning has recently shown promise as an alternative to finetuning, with the major advantage that the same model can be served for multiple tasks. However, this shifts the computational burden from training-time to inference-time, making deployment of many-shot ICL challenging to justify in-practice. This cost is further increased if a custom demonstration set is retrieved for each inference example. We present \ours, a training-free framework for retrieval-based many-shot in-context learning. By combining carefully designed block-sparse attention and retrieval of cached \textit{groups} of demonstrations, we achieve comparable per-example latency to finetuning while maintaining  on average >95\% of the best method's accuracy across strong ICL and finetuning baselines. We hope that this will further enable the deployment of many-shot ICL at scale.%
\footnote{Data and code are available at \url{https://github.com/millix19/dbsa}}
\end{abstract}

\section{Introduction}

When adapting large language models (LLMs) to a specific task, practitioners typically use finetuning or in-context learning (ICL). ICL has the advantage of convenience— it requires no parameter updates, adapts easily to new tasks and datasets, and does not require serving task-specific models in a production setting. In the standard few-shot setting, ICL typically underperforms finetuning \cite{mosbach-etal-2023-shot} and requires a careful demonstration curation process \cite{dong2024survey}. 

Recently, many-shot ICL—which uses thousands of demonstrations—shows performance comparable to finetuning, and even surpasses it for various tasks \cite{bertsch2024LC-ICL,agarwal2024many}. 
However, scaling ICL to long contexts introduces a new computational tradeoff: many-shot ICL shifts the computational burden from fine-tuning to inference, creating significant efficiency challenges. If the same fixed set of demonstrations is used for all requests, some computation can be cached; however, performance is higher when a custom set of demonstrations is retrieved. Processing thousands of demonstrations per request is orders of magnitude more expensive than zero-shot inference with a fine-tuned model, making many-shot ICL impractical for high-throughput applications.


\begin{figure}
    \centering
    \includegraphics[width=0.9\linewidth]{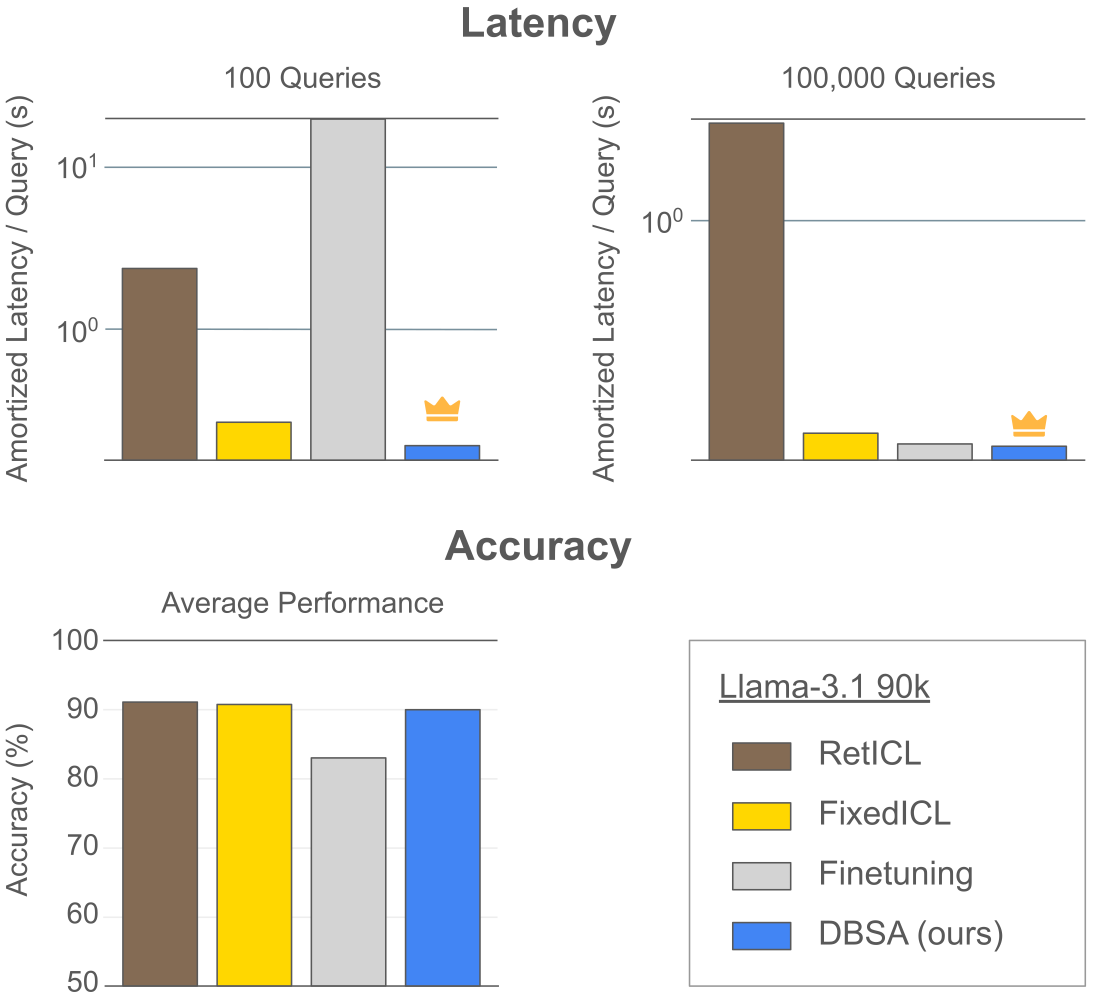}
    \caption{\abbrours maintains high accuracy while achieving the best overall efficiency when compared to many-shot ICL baselines and finetuning, even under high request volumes.}
    \label{fig:first fig}
\end{figure}

In this paper, we propose \ours (\abbrours), a training-free inference framework that minimizes many-shot ICL latency while maintaining >95\% of the best accuracy on average across all baselines, including many-shot ICL, retrieval ICL, and finetuning. We introduce key optimizations to both demonstration pre-encoding and inference. 

During pre-encoding, we apply a structured block-sparse streaming attention pattern \cite{xiao2024efficient}, where each demonstration attends only to a fixed number of others and a global attention sink. 
Then, during inference, we integrate retrieval ICL with KV cache reuse, dynamically selecting \textit{groups} of relevant demonstrations that were pre-encoded together for each test query. The test query attends only to the KV cache of selected groups to save computation.

\begin{figure}
    \centering
    \includegraphics[width=0.9\linewidth]{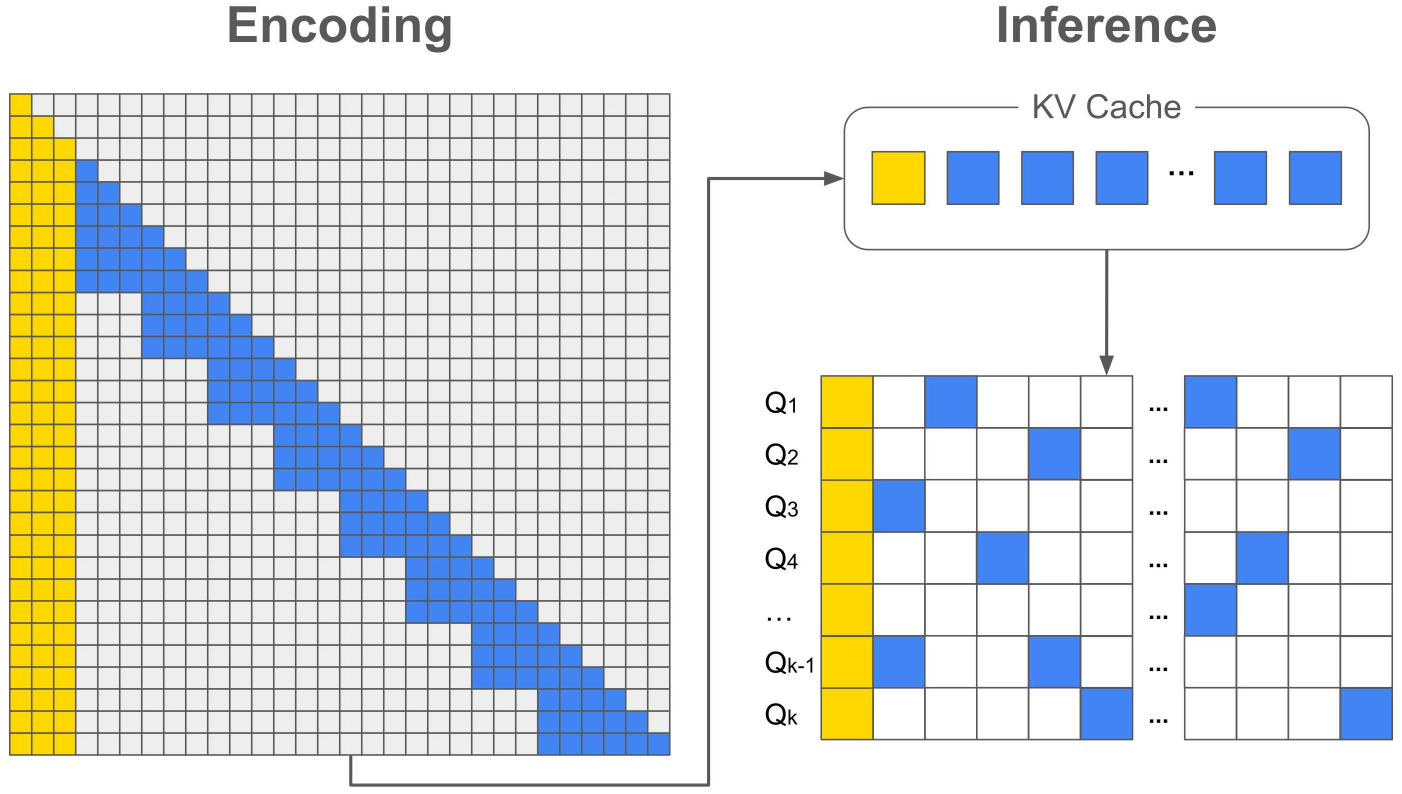}
    \caption{\abbrours pre-encodes the many-shot demonstration pool with block-sparse attention, and dynamically selects relevant KV chunks during inference for each test query.}
    \label{fig:second fig}
\end{figure}

Unlike standard key-value (KV) cache reuse, with \abbrours the encoding time for new demonstrations remains constant as the demonstration pool grows, a benefit for practical scenarios where the demonstration pool may change over time. 
Importantly, this mechanism works without any additional training or optimization. And unlike standard retrieval ICL, \abbrours reuses a pre-encoded cache, combining the efficiency benefits of pre-encoding a fixed demonstration set with the performance benefits of retrieving custom demonstrations.




We evaluate \abbrours on five datasets using Llama-2 and Llama-3.1 models with up to 90k tokens of in-context demonstrations. 
Our second-stage inference speed up is proportional to the retrieval ratio -- at a 30\% retrieval ratio, our method achieves more than 2× speedup over cached many-shot ICL; when the selected KV cache is within 30k tokens, \abbrours reaches inference latencies comparable to a finetuned Llama-3.1, with >1000x less setup time and better accuracy. 
Our results 
demonstrate many-shot ICL can be the most competitive solution in practice, even in scenarios with large request volumes.

\section{Methodology}


\ours treats many-shot ICL as a two-stage process. First, there is a one-time setup cost to encode the many-shot demonstrations; and second, a dynamic KV-cache is selected for each answer generation.

\subsection{Stage 1: Many-Shot Encoding}

In the first stage, we aim to encode the full demonstration set with minimal cost. Note that the full demonstration set may exceed the maximum context size of our model, even when using a long-context model (e.g. if there are tens of thousands of training examples). 

Prior research  \cite{bertsch2024LC-ICL, acharya2024starattention, Lu2024TurboRAGAR} suggests that letting each example attend to some number of other examples is beneficial, but attending to the entire demonstration pool is often unnecessary to achieve high performance. We use a structured sparse attention pattern for efficiency.

Given a set of demonstrations \(D = \{d_1, d_2, ..., d_n\}\), where each demonstration \(d_i\) consists of a query-answer pair \((q_i, a_i)\), we first partition \(D\) into \(\frac{n}{k}\) blocks: \(D = [b_1, b_2, ..., b_{n/k}]\), where each block \(b_i\) contains \(k\) demonstrations. We encode \(D\) using a block-wise streaming attention mechanism, where each block \(b_i\) attends to three key components: (1)  an anchor block \(b_1\) , (2) a fixed number (\(j\)) of previous blocks \(\{b_{i-j}, ..., b_{i-1}\}\), and (3) itself using standard causal attention. 
Because not all examples attend to each other, the demonstrations do not all need to be encoded in the same forward pass or even at the same time; thus, new demonstrations can be added to the demonstration pool at any point by encoding an additional block with this attention pattern. 

Following the same approach as StreamingLLM \cite{xiao2024efficient}, we encode \(D\)  using sequential position ids \([0, ... n-1]\) and cache the KV states for each block before applying the rotary position transformation. However,  unlike StreamingLLM, we retain all KV cache entries, as each test query benefits from a distinct set of demonstrations. 

The many-shot encoding stage is implemented using Flex Attention \cite{dong2024flexattention}, which skips the computation for masked blocks and achieves a proportional performance speedup compared to FlashAttention-v2 \cite{dao2024flashattention2}.

\subsection{Stage 2: Dynamic Demonstration Selection and Answer Generation}

Prior research has demonstrated that retrieval ICL can achieve strong performance using only a subset of the available demonstration pool \cite{luo2024retrievedICL}. However, achieving peak performance still requires hundreds of examples, which remains computationally expensive with a naive implementation that re-encodes the context for every test query. Our method reduces this inference overhead by reusing the precomputed KV caches from Stage 1.


Given a test query $q^*$, a retrieval method is applied to select a subset of demonstration blocks $D' \subset D$, where $D' = \{b'_1, b'_2, ..., b'_m\}$, $m < n$. The anchor block $b_1$ is always included as $b'_1$ to serve as the attention sink.  We concatenate the KV caches of the selected demonstrations, and re-apply relative positional encoding based on new in-order position IDs ranging from \( [0, |D'| - 1] \). Finally, we encode $q^*$ with full attention to the selected KV cache, and generate the answer $a^*$ autoregressively.

\abbrours provides the flexibility to plugin any retrieval method of choice, including text-based similarity, cosine similarity, or diversity-focused retrievers \cite{dong2024survey, luo2024retrievedICL}. The choice of retrieval method also presents an efficiency-performance tradeoff; for this work, we use the inexpensive and relatively strong BM25 retriever \cite{robertson2009probabilistic}. 


\section{Experimental Setup}
\textbf{Datasets.} We evaluate our method on 5 language classification datasets—TREC \citep{hovy-etal-2001-toward}, TREC-fine  \citep{hovy-etal-2001-toward}, NLU \citep{nlu}, Banking-77  \citep{casanueva-etal-2020-efficient}, and Clinic-150 \citep{clinic150}. These datasets span a diverse range of domains and label spaces, are commonly used in prior work in ICL \cite{han2022prototypical,ratner2023parallel}, and have been explored in the many-shot ICL setting \cite{bertsch2024LC-ICL,yen2024helmetevaluatelongcontextlanguage}. 

We consider two context lengths, 30k and 90k, and construct the demonstration pool $D$ by randomly selecting the maximum number of demonstrations that fit in-context; for these classification datasets, which have short input/output pairs, thousands of demonstrations fit in 90k context. Table \ref{tab:dataset-stats} provides summary statistics for each dataset. 

\begin{table}[ht]
\centering
\small
\setlength{\tabcolsep}{4pt} 
\renewcommand{\arraystretch}{1.5}
\resizebox{\columnwidth}{!}{ 
\begin{tabular}{lcccc}
\toprule
\textbf{Dataset} & \textbf{Domain} & \textbf{\# Labels} & \textbf{\makecell[c]{\# Demo \\ (30k Context)}} & \textbf{\makecell[c]{\# Demo \\ (90k Context)}} \\
\midrule
TREC        & Questions      & 6    & 1050  & 3150  \\
TREC Fine   & Questions      & 50   & 1000  & 3000  \\
NLU         & Conversational & 68   & 1150  & 3450  \\
Banking-77  & Financial      & 77   & 800   & 2400  \\
Clinic-150  & Multiple       & 151  & 1000  & 3000  \\
\bottomrule
\end{tabular}
}
\caption{Dataset statistics.}
\label{tab:dataset-stats}
\end{table}

\textbf{Models.} We consider 2 models in the Llama model series for evaluation. Llama-2-7B (32k) \citep{TogetherAI_2023}  is the base Llama-2-7B model finetuned with an extended 32k context window. We also consider the more recent Llama-3.1-8B \cite{meta2024llama3.1}, which supports a 128k context window and incorporates Grouped-Query Attention (GQA) for efficient long-context inference. 

\textbf{Baselines.} We compare our method against three baselines: \fixedicl,  \reticl, and \ft. \textit{\fixedicl} uses the entire 30k or 90k demonstration set as context for every test query. We consider the efficient implementation where the context is encoded once and its KV cache reused across test queries. \textit{\reticl} (RetICL) dynamically selects a subset of relevant demonstrations from the 30k or 90k demonstration pool for each test query. In this scenario, a distinct context has to be encoded for each test query at inference time. As retrieval strategy is a flexible choice, we use BM25 retriever \cite{bassani2023retriv,robertson2009probabilistic} for both this baseline and our method to ensure a fair comparison. 
 \textit{Finetuning} uses LoRA \cite{hu2021lora} with lora rank 8.  We choose the best setup and hyperparameters based on ablation results in \citet{bertsch2024LC-ICL}. For additional details, see Appendix~\ref{appendix:finetune}. For all ICL methods, we use constrained decoding to output only valid labels. For finetuning, we finetune with a classification head. 


\textbf{Evaluation.} 
Following prior work, we randomly sample 250 test examples from each dataset. We evaluate both accuracy and efficiency metrics, averaged over 10 runs. 
We compare our method to the baseline using FlashAttention-2 \cite{dao2024flashattention2}. All methods with 30k context are run on a single L40S GPU (48GB) and all methods with 90k context are run on a single A100 GPU (80GB).

\textbf{Configuration.} 
We use 50 examples per block, with demonstration blocks formed through random grouping. Across all context lengths, we designate the first block as the attention sink and allow each block to attend to the two preceding blocks as locally attended context. For both the Retrieval ICL baseline and our method, we retrieve 30\% of the total demonstration pool for each test query. This retrieval ratio is used for all datasets.

\section{Main Results}
\label{sec:main results}

We compare \abbrours to the baselines along two dimensions: efficiency and accuracy. 

\begin{figure}[t]
    \centering
    \begin{subfigure}{0.48\linewidth}
        \centering
        \includegraphics[width=\linewidth]{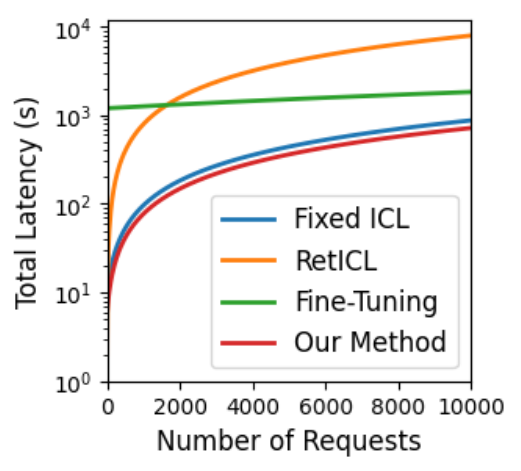}
        \caption{Llama-3.1-8B 90k}
        \label{fig:latency-llama3}
    \end{subfigure}
    \hspace{1mm} 
    \begin{subfigure}{0.47\linewidth}
        \centering
        \includegraphics[width=\linewidth]{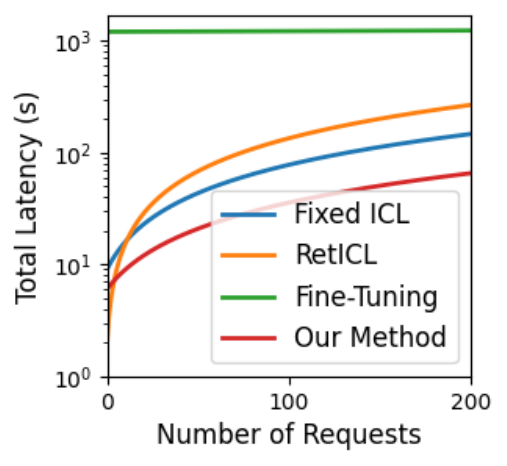}
        \caption{Llama-2-7B 30k}
        \label{fig:latency-llama2}
    \end{subfigure}
    
    \caption{The total inference latency for each method, including setup and per-request inference latency. We use inference batch size 1.}
    \label{fig:latency-figures}
\end{figure}

\subsection{Efficiency Comparison}


Efficiency is the primary motivation behind our method, which aims to close the inference latency gap between many-shot ICL and zero-shot inference with a finetuned model. To evaluate efficiency, we consider setup overhead, inference latency, and storage cost of each method. 

Figure \ref{fig:latency-figures} and \ref{fig:first fig} shows that \abbrours is the most efficient overall, even with more than 100,000 requests. The latency comparison in Table \ref{tab:latency-comparison} breaks down the efficiency gains, showing that \abbrours can bring inference costs close to those of zero-shot inference while maintaining a significantly lower setup time than fine-tuning.

\begin{table}[ht]
\centering
\small
\setlength{\tabcolsep}{6pt} 
\renewcommand{\arraystretch}{1.4}
\resizebox{\columnwidth}{!}{ 
\begin{tabular}{lcc}
\toprule
\textbf{Method} & \textbf{\makecell[c]{Setup Time \\ (Relative)}} & \textbf{\makecell[c]{Inference Latency \\ (Relative)}} \\
\midrule
\multicolumn{3}{l}{\textbf{30k Context Length w/ Llama-2-7B w/ single L40S gpu}} \\
\midrule
RetICL (no cache)  & \textbf{1x}  & 1x  \\
Fixed ICL (cached) & 4.5x  & 0.51x  \\
Finetuning (LoRA)       & > 600x  & \textbf{0.12x}  \\
\abbrours (ours)         & \underline{3x}  & \underline{0.22x}  \\

\midrule
\multicolumn{3}{l}{\textbf{30k Context Length w/ Llama-3.1-8B w/ single L40S gpu}} \\
\midrule
RetICL (no cache)  & \textbf{1x}  & 1x  \\
Fixed ICL (cached) & 5x  & 0.11x  \\
Finetuning (LoRA)       & > 600x  & \textbf{0.08x}  \\
\abbrours (ours)         & \underline{3x}  & \underline{0.10x}  \\

\midrule
\multicolumn{3}{l}{\textbf{90k Context Length w/ Llama-3.1-8B w/ single A100 gpu}} \\
\midrule
RetICL (no cache)  & \textbf{1x}  & 1x  \\
Fixed ICL (cached) & 6.5x  & 0.06x  \\
Finetuning (LoRA)     & > 1500x  & \textbf{0.046x}  \\
\abbrours (ours)         & \underline{4x}  & \underline{0.053x}  \\

\bottomrule
\end{tabular}
}
\caption{\label{tab:latency-comparison} Relative latency of different methods compared to RetICL baseline. Best method is bolded and second best is underlined.  Since the absolute numbers depend on the computational environment, we present the relative speedups here; the raw numbers for our hardware setup are available in Appendix~\ref{appendix:efficiency}}
\end{table}

\textbf{Setup overhead.} Retrieval ICL incurs the lowest setup time (in our case, only the time to construct the BM25 retriever index). Fixed ICL is more expensive at setup time, as setup requires encoding the full 30k or 90k demonstration set with full attention. Fine-tuning has the highest setup cost, as it requires training the model with LoRA for approximately 30 epochs to reach high performance. In practice, finetuning took about 30 minutes with a 30k demo pool, though here we exclude the time for evaluation steps in our reported time. 

\abbrours involves both building the retriever index and encoding the full demonstration pool. However, it still achieves a speedup over Fixed ICL thanks to the structured sparse attention. While setup cost is a one-time fixed cost with a static demonstration pool, real-world demonstrations are often non-static. In these cases, the speedup from sparse encoding becomes particularly valuable.

\textbf{Inference latency}, or the per-example cost, is key to the practicality of deployment. Compared to Fixed ICL, \abbrours lowers latency by attending to a reduced KV cache context length. Theoretically, inference latency grows linearly with the size of the KV cache, a trend empirically shown in \citet{agarwal2024many}. Therefore, smaller retrieval ratios can lead to further improvements.

While Retrieval ICL without cached encoding is always the most expensive in terms of latency per request, the latency gap between cached ICL and finetuning narrows significantly when using Llama-3.1 compared to Llama-2, making cached many-shot ICL increasingly competitive. One major factor contributing to this speedup is Grouped-Query Attention (GQA), which reduces the computational cost of handling long sequences. All ICL methods benefit from this speedup; however, as fixed ICL uses the most demonstrations in-context, it is disproportionately impacted. Despite this, \abbrours is still slightly faster per-example. 

In Figure \ref{fig:first fig} and \ref{fig:latency-figures}, we consider an alternate metric: the total compute time necessary to accomplish the task, amortized over the total number of inference requests. This includes both setup overhead and the per-example latency. \abbrours remains more efficient than finetuning even when more than 100,000 requests are made.

\textbf{Storage Cost Considerations.} Beyond computational efficiency, we also evaluate the storage cost. For Llama-3.1-8B, the KV cache requires 0.125 MiB per token, which is 3.7GB for a 30k demonstration pool and 11.1GB for 90k. While manageable for moderate demonstration sizes, this storage demand may require offloading or re-encoding for larger demonstration pools or multi-task settings. Note that this is a shared cost across all many-shot ICL methods with long context. 

For a single task, the storage cost for LoRA fine-tuning is significantly smaller. With Llama-3.1-8B model and LoRA rank 8, the storage is 0.01GB per task. However, in settings where hundreds or thousands of tasks must be supported—such as user-specific tasks—the cumulative storage cost of fine-tuning also becomes substantial. Furthermore, task-switching and task addition introduce additional overhead \cite{xia2024efficient,sheng2023slora,gabrielsson2024compress}. In these cases, the fast setup and flexibility of ICL provide a strong advantage over fine-tuning-- if storage cost is a limited factor, demonstration caches can be recomputed when needed instead of stored indefinitely. 


\begin{figure}
    \centering
    \includegraphics[width=0.85\linewidth]{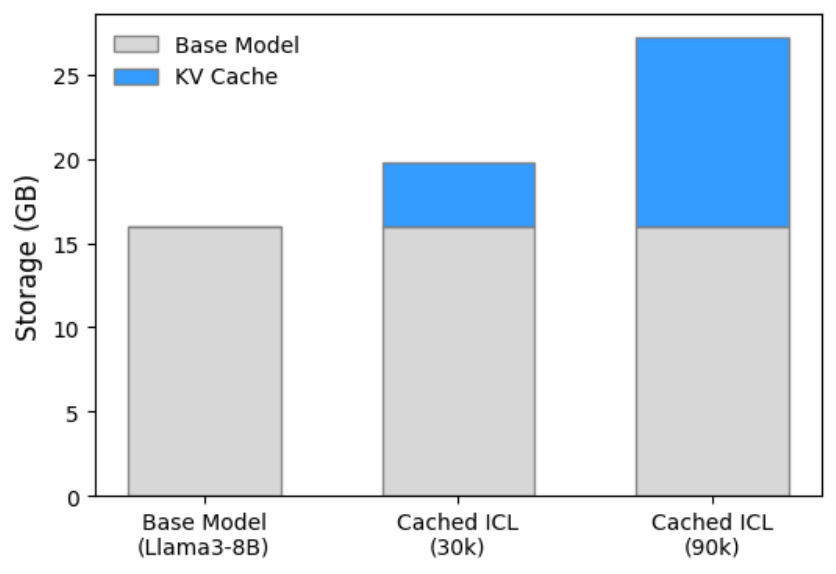}
    \caption{Storage cost for Cached many-shot ICL, which encompass \fixedicl and \abbrours}
    \label{fig:storage-compare}
\end{figure}


\subsection{Accuracy Comparison}

\begin{table}[ht]
\centering
\small
\setlength{\tabcolsep}{2pt} 
\renewcommand{\arraystretch}{1.3}
\resizebox{\columnwidth}{!}{ 
\begin{tabular}{lcccc}
\toprule
\textbf{Dataset} & \textbf{Fixed ICL} & \textbf{Ret ICL} & \textbf{Fine-Tuning} & \textbf{\abbrours (ours)} \\
\midrule
\multicolumn{5}{l}{\textbf{30k Context Length w/ Llama-2-7B }} \\
TREC        & 0.93  & 0.92  & \textbf{0.95}  & 0.91 (\textcolor{blue}{96\%}) \\
TREC Fine   & 0.76  & \textbf{0.79}  & 0.77  & 0.77 (\textcolor{blue}{97\%}) \\
Banking77   & 0.81  & \textbf{0.84}  & 0.80  & 0.80 (\textcolor{blue}{95\%}) \\
Clinic      & \textbf{0.86}  & 0.83  & 0.60  & 0.80 (\textcolor{blue}{93\%}) \\
NLU         & 0.85  & \textbf{0.86}  & 0.74  & 0.84 (\textcolor{blue}{98\%}) \\
Average    & 0.84 & 0.85 & 0.77 & 0.84 (\textcolor{blue}{97\%}) \\

\midrule
\multicolumn{5}{l}{\textbf{90k Context Length w/ Llama-3.1-8B}} \\
TREC        & 0.96  & 0.95  & \textbf{0.96}  & 0.95 (\textcolor{blue}{99\%}) \\
TREC Fine   & 0.88  & \textbf{0.89}  & 0.83  & 0.88 (\textcolor{blue}{99\%}) \\
Banking77   & \textbf{0.91}  & 0.90  & 0.81  & 0.89 (\textcolor{blue}{98\%}) \\
Clinic      & \textbf{0.92}  & 0.91  & 0.74  & 0.90 (\textcolor{blue}{98\%}) \\
NLU         & 0.89  & \textbf{0.90}  & 0.82  & 0.88 (\textcolor{blue}{98\%}) \\
Average    & 0.91 & 0.91 & 0.83 & 0.90 (\textcolor{blue}{99\%}) \\

\bottomrule
\end{tabular}
}
\caption{\label{tab:accuracy-comparison} Accuracy comparison. Bolded is the max accuracy for each dataset. Blue percentage is how \abbrours compares to the best accuracy.}
\end{table}

Table \ref{tab:accuracy-comparison} show that \abbrours consistently achieves high accuracy across datasets, closely matching the best baseline method. This is particularly evident at 90k demonstration context, which is also where accuracy of these datasets saturate, making it the most relevant comparison point. 

The success of our approach is largely due to the effectiveness of \reticl, which is often the best performing baseline. We set a fixed retrieval ratio of 30\% across all datasets and context lengths. Even without optimizing the retrieval ratio, \reticl achieves accuracy comparable to \fixedicl using the entire demonstration pool. Prior work suggests that \reticl can surpass \fixedicl when the retrieval ratio is tuned optimally \cite{luo2024retrievedICL, bertsch2024LC-ICL}. However, determining the ideal retrieval ratio is non-trivial-- performance initially improves with more demonstrations, but eventually degrades due to noise from less relevant demonstrations. While we use a fixed retrieval ratio, our framework supports adaptive selection strategies to optimize retrieval for each query. Given that our method’s accuracy ceiling is constrained by the upper bound of Retrieval ICL, further improvements in retrieval quality would benefit our method as well.

At 30k context length, we observe a more noticeable accuracy gap between \abbrours and \reticl. This comes from two key design choices. First, we always include the first 50-example block as the attention sink, which may not always be relevant to the test query. Second, we group blocks randomly and treat each block as a unified text chunk when computing relevant scores. This raises the risk of missing highly relevant examples. In Section \ref{sec:group} and Section \ref{sec:block vs ind}, we analyze different block grouping strategies and discuss potential refinements that can balance retrieval relevance and diversity more effectively, and why block-level selection is still necessary.


\begin{table}[ht]
\centering
\small
\setlength{\tabcolsep}{4pt} 
\renewcommand{\arraystretch}{1.4}
\resizebox{\columnwidth}{!}{ 
\begin{tabular}{lccc}
\toprule
\textbf{Dataset} & \textbf{Fine-Tuning (90k)} & \textbf{Fine-Tuning (All)} & \textbf{\abbrours (90k)} \\
\midrule
TREC        & 0.96  & 0.96  & 0.95  \\
TREC Fine   & 0.83  & 0.89  & 0.88  \\
Banking77   & 0.81  & 0.91  & 0.89  \\
Clinic      & 0.74  & 0.89  & 0.90  \\
NLU         & 0.82  & 0.87  & 0.88  \\
\midrule
\textbf{Avg} & 0.832 & 0.904 & 0.900 \\
\bottomrule
\end{tabular}
}
\caption{\label{tab:finetune-comparison} Accuracy comparison of finetuning with 90k demonstration pool, or using the full training dataset.}
\end{table}

We observe that finetuning underperforms significantly compared to many-shot ICL, even with 90k demonstration pool (3k demonstrations). This aligns with previous findings \cite{bertsch2024LC-ICL}, which show that finetuning requires a larger number of demonstrations to achieve competitive or higher performance. However, even when trained on the maximum available data (5k to 20k demonstrations for our datasets), finetuning fails to substantially surpass the best accuracy achieved by many-shot ICL, as shown in Table \ref{tab:finetune-comparison}. 

\section{Key Insights and Ablations}

We conduct ablations to analyze the impact of various design decisions in our framework.


\subsection{Sparse Attention for Many-shot ICL}
\label{sec:sparse attention}


\begin{figure}[ht]
    \centering
    \begin{subfigure}{0.44\linewidth}
        \centering
        \includegraphics[width=\linewidth]{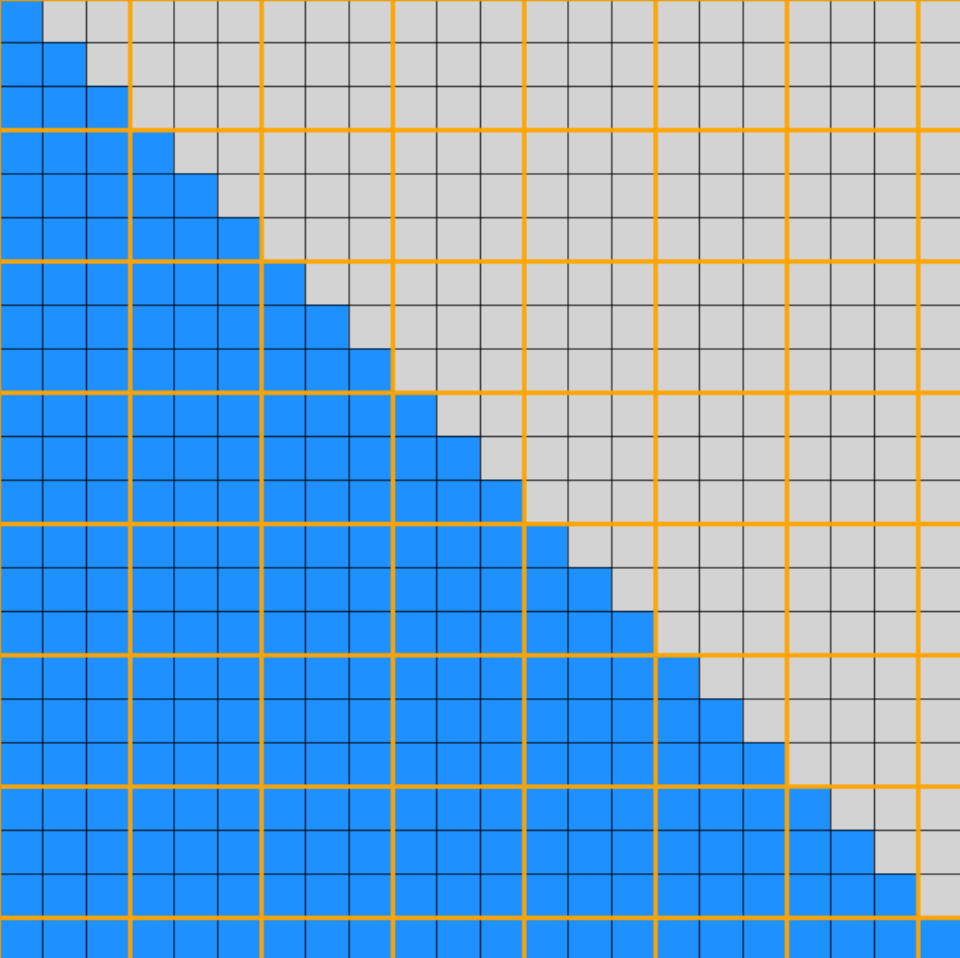}
        \caption{Full Attention}
        \label{fig:attn-full}
    \end{subfigure}
    \hfill
    \begin{subfigure}{0.44\linewidth}
        \centering
        \includegraphics[width=\linewidth]{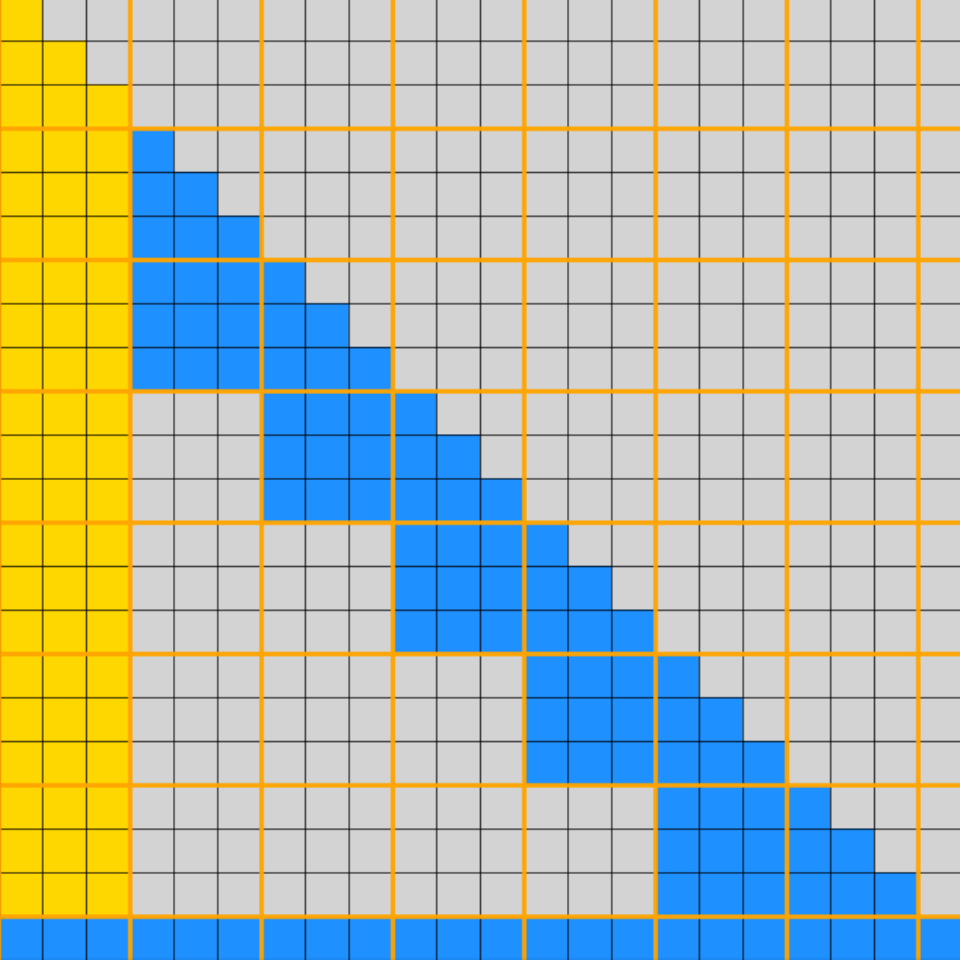}
        \caption{Sink + Prev + Self}
        \label{fig:attn-sink-self-prev}
    \end{subfigure}

    \vspace{0.5cm} 

    \begin{subfigure}{0.44\linewidth}
        \centering
        \includegraphics[width=\linewidth]{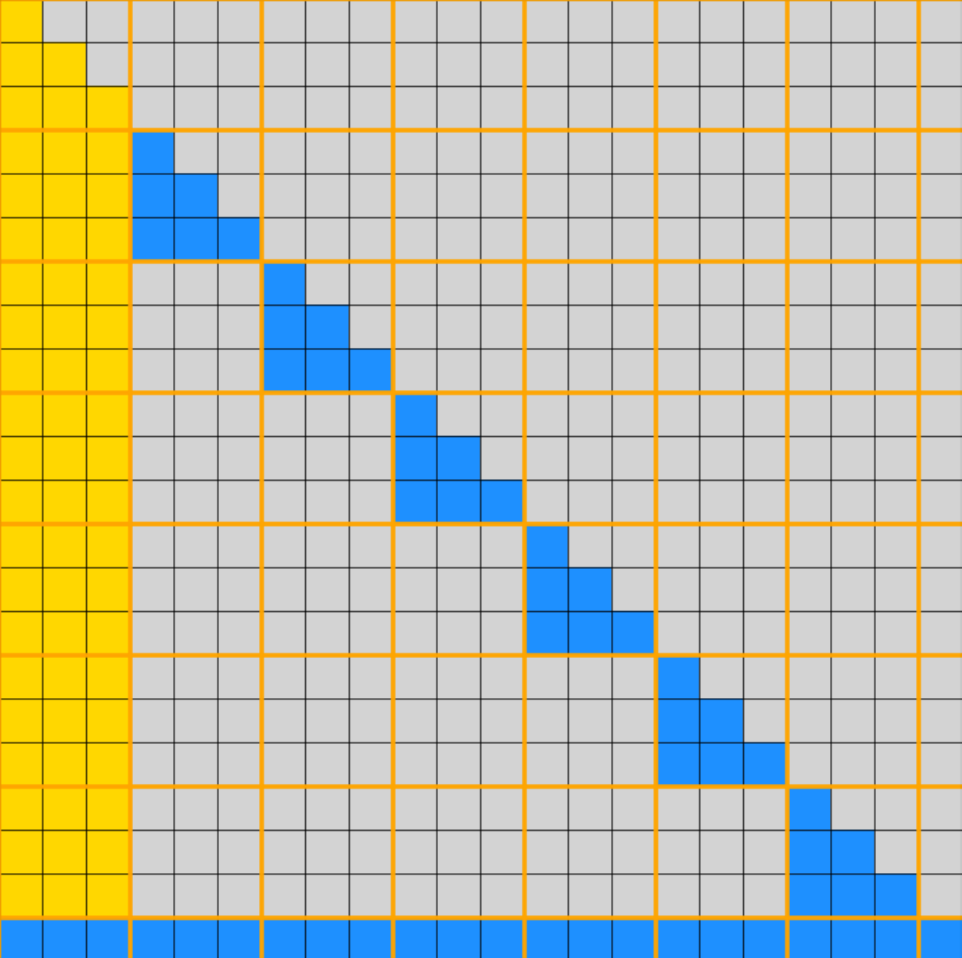}
        \caption{Sink + Self}
        \label{fig:attn-self-sink}
    \end{subfigure}
    \hfill
    \begin{subfigure}{0.44\linewidth}
        \centering
        \includegraphics[width=\linewidth]{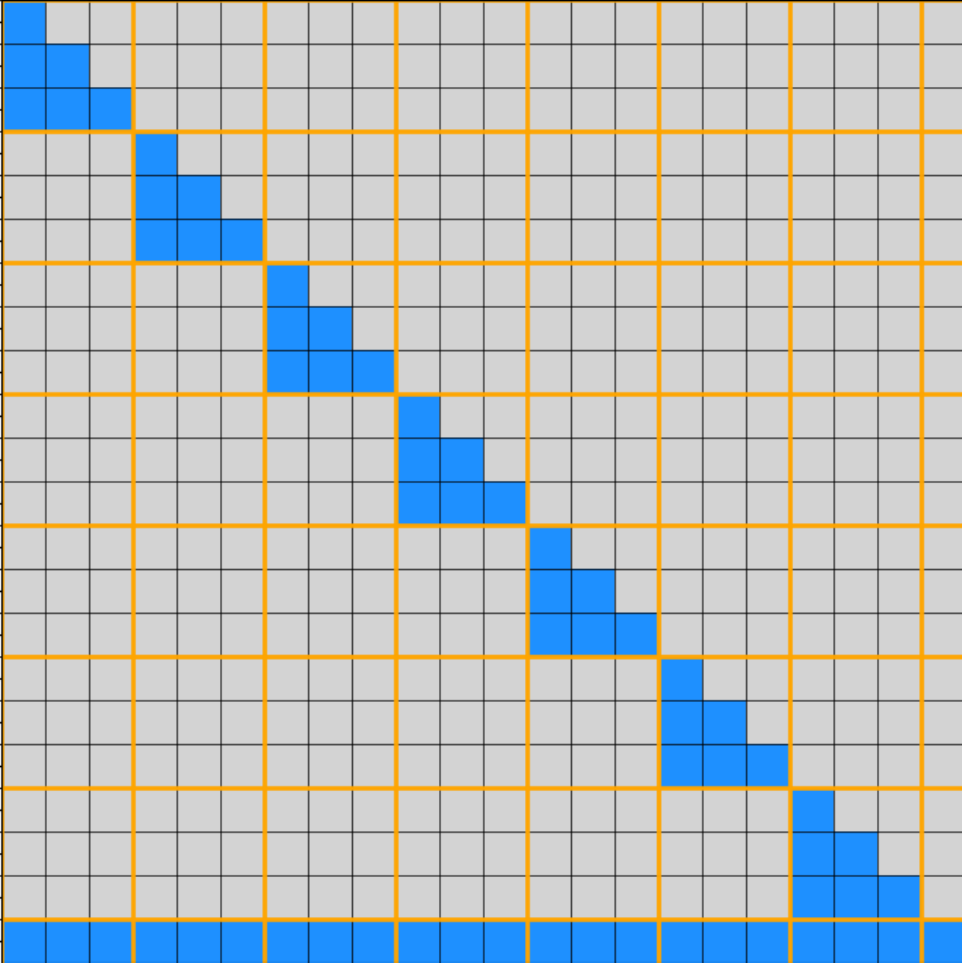}
        \caption{Self}
        \label{fig:attn-self}
    \end{subfigure}

    \caption{Visualization of block-sparse attention mechanisms for many-shot ICL in section \ref{sec:sparse attention}.
    }
    \label{fig:attn-comparison}
\end{figure}

The success of our sparse encoding approach shows that not all interactions within the demonstration set are necessary by default. However, sparse attention paradigms have been extensively explored since the introduction of self-attention, with varying applicability—some operate at inference time, while others require retraining for adaptation. Here, we ablate over design decisions in the sparse attention to determine what makes this pattern effective without additional training.

While sparsity alone is not always optimal, block-wise sparsity considers I/O and memory efficiency, making it practical for real-world usage \cite{dong2024flexattention,dao2024flashattention2, guo2024blocksparse}. In this section, we compare variations of structured blockwise sparse attentions for many-shot encoding, illustrated in Figure \ref{fig:attn-comparison}:

\begin{itemize}
    \setlength{\itemsep}{0pt}  
    \setlength{\parskip}{0pt}  
    \item \textbf{Full Attention}: Each token attends to all other tokens.
    \item \textbf{Sink + Prev + Self}: Each block attends to an \textit{anchor block}, a few \textit{preceding blocks}, and \textit{itself}. This is the pattern applied in \abbrours.
    \item \textbf{Sink + Self}: Each block attends to a \textit{sink block} and \textit{itself} only.
    \item \textbf{Self-Only}: Each block attends \textit{only to itself}.
\end{itemize}

These mechanisms have been applied to some extent in prior work, and are largely inspired by StreamingLLM \cite{xiao2024efficient}. To evaluate these patterns, we allow test queries to attend to the full encoded context and compare the accuracies.

\begin{table}[ht]
\centering
\setlength{\tabcolsep}{7pt} 
\renewcommand{\arraystretch}{1.2}
\resizebox{\columnwidth}{!}{%
\begin{tabular}{lcccc}
\toprule
\textbf{Dataset} & \textbf{Full} & \textbf{sink+prev+self} & \textbf{sink+self} & \textbf{self} \\
\midrule
Banking77  & 0.81 & 0.80 & 0.34 & 0.02 \\
Clinic150   & 0.85 & 0.80 & 0.33 & 0.02 \\
NLU      & 0.86 & 0.82 & 0.27 & 0.06 \\
TREC & 0.93 & 0.91 & 0.24 & 0.21 \\
TREC Fine & 0.76 & 0.76 & 0.17 & 0.151 \\
\midrule
\textbf{Avg} & 0.84 & 0.82 & 0.27 & 0.09 \\
\bottomrule
\end{tabular}%
}
\caption{\label{tab:attention-patterns-comparison}
Comparison of accuracies using different sparse attention patterns, using Llama-2, 30k demo size, and block size of 50 demos.
}
\end{table}
Table \ref{tab:attention-patterns-comparison} shows that both an attention sink and local context connections are necessary for sparse attention to achieve performance comparable to full attention without additional training.

These findings align with previous research that applied similar patterns. The "Sink + Self" pattern has been used in StarAttention \cite{acharya2024starattention} for training-free inference speedup, but the results indicate that it performs well only when the number of blocks remains small (fewer than four). Other works also adopted this pattern \cite{Lu2024TurboRAGAR} \cite{sun2024blockattentionefficientrag}, but they required finetuning for adaption. Meanwhile, the "Sink + Prev + Self" streaming attention \cite{xiao2024efficient}, commonly used in KV eviction strategies \cite{xiao2024duo} \cite{li2024snapkv}, performs surprisingly well in our many-shot ICL setting. It remains scalable as demo size increase, as the same block size cause minimum degration for both 30k and 90k setting. At 90k context, the attention mask is over 90\% sparse while maintaining strong performance.

The scalability means that new demonstrations simply needs to attend to a fixed length of context as demo pool grows. Additionally, when KV cache storage is limited, applying this sparsity improves inference efficiency even in no-cache settings.



\subsection{Block vs. Individual Demonstration Selection}
\label{sec:block vs ind}




In our framework, we group demonstrations into blocks for encoding and KV selection, rather than selecting individual examples like the \reticl baseline. In this section, we compare the two and separate the effects of using different selected examples and re-using KV cache segments. 

Consider a set of demonstrations [1, 2, 3, 4, 5, 6, 7, 8] with sparse encoding, where [1] serves as the attention sink and each demonstration attends to itself and one preceding example (e.g., 5 attends to [1, 4, 5]). If we use example-level retrieval and retrieve [3, 6], 6 would not have attended to 3 during encoding. In contrast, if we select [3, 4] as a block, then it would be a contiguous KV cache segment. We hypothesize that block-level selection could better preserve intra-context relationships during encoding, and thus have better performance. 

To test this hypothesis, we run block-level and example-level selection with \abbrours, and see how each compare with the standard inference without sparse encoding and KV cache reuse. We make sure that the \abbrours and standard inference use the same examples in context to isolate the effect of segmented KV cache reuse.Results in Table \ref{tab:block-vs-example} show that example-level selection with \abbrours is also effective, but block-level selection has slightly higher accuracy; it also has faster retrieval time and memory efficiency when handling KV caches. 



\begin{table}[ht]
\centering
\small
\setlength{\tabcolsep}{3pt} 
\renewcommand{\arraystretch}{1.3} 
\resizebox{\columnwidth}{!}{%
\begin{tabular}{lcccccc}
\toprule
\textbf{Setting} & \textbf{example} & \textbf{\makecell[c]{example \\ (\abbrours)}} & \textbf{diff} & \textbf{block} & \textbf{\makecell[c]{block \\ (\abbrours)}} & \textbf{diff} \\
\midrule
30k - Llama-2  & 0.82 & 0.79 & \textbf{0.03} & 0.79 & 0.79 & \textbf{0.00} \\
30k - Llama-3.1  & 0.86 & 0.82 & \textbf{0.04} & 0.84 & 0.82 & \textbf{0.02} \\
90k - Llama-3.1  & 0.90 & 0.86 & \textbf{0.04} & 0.89 & 0.88 & \textbf{0.01} \\
\bottomrule
\end{tabular}%
}
\caption{\label{tab:block-vs-example} Accuracy comparison of block-level and example-level retrieval, using standard inference vs. \abbrours with sparse encoding and KV reuse. All use a 10\% retrieval ratio.}
\end{table}

\vspace{-10pt}
\subsection{Block Grouping}
\label{sec:group}

In our main experiments, we randomly group the demonstration pool into blocks before encoding and retrieval. However, the way demonstrations are grouped may impact accuracy. In this section, we explore whether alternative grouping strategies lead to improved performance. 

We evaluate three grouping strategies: (1) random grouping, (2) k-means clustering based on BM25 similarity, and (3) starting with BM25 clustering, we swap 10\% examples between clusters to introduce diversity within each block. 

Table ~\ref{tab:block-grouping} shows that introducing some diversity within blocks improves performance compared to strictly clustered examples. Interestingly, random grouping remains competitive, sometimes outperforming clustering-based methods. A possibility is that random grouping naturally leads to diversity within a block, thus achieving the same goals.



\begin{table}[ht]
\centering
\small
\setlength{\tabcolsep}{5pt} 
\renewcommand{\arraystretch}{1.3} 
\resizebox{\columnwidth}{!}{%
\begin{tabular}{l p{2cm} p{2cm} p{2.2cm}}
\toprule
\textbf{Dataset} & \textbf{Random} & \textbf{Clustered} & \textbf{\makecell[c]{Clustered \\ (w/ diversity)}} \\
\midrule
Banking77 & \hspace{5pt}0.820 & \hspace{8pt}0.696$^*$ & \hspace{19pt}0.740$^*$ \\
Clinic    & \hspace{5pt}0.796 & \hspace{8pt}0.792 & \hspace{19pt}0.796 \\
NLU       & \hspace{5pt}0.840 & \hspace{8pt}0.792$^*$ & \hspace{19pt}0.832 \\
TREC      & \hspace{5pt}0.912 & \hspace{8pt}0.910 & \hspace{19pt}0.916 \\
TREC Fine & \hspace{5pt}0.766 & \hspace{8pt}0.692$^*$ & \hspace{19pt}0.764 \\
\bottomrule
Avg       & \hspace{5pt}0.827 & \hspace{8pt}0.764 & \hspace{19pt}0.810 \\
\midrule
\end{tabular}
}
\caption{\label{tab:block-grouping} 
Comparison of block grouping strategies with \abbrours. * denotes significance at p<0.05 using T-test against random grouping.
}
\end{table}

\vspace{-13pt}  
\subsection{Dynamic Block Ordering}
The "Lost in the Middle" phenomenon suggests that long-context models are sensitive to position of relevant information, with content placed at the beginning or end of the context having a stronger influence \citep{liu2024lost,jin2024longcontextllmsmeetrag}. 

We evaluate whether, within our framework, the KV cache of demonstrations is robust to reordering and whether reordering strategies can be leveraged to enhance performance. Using BM25 relevance scores, we rank blocks for each test query and compare three reordering approaches: (1) maintaining the original encoded order, (2) re-arranging blocks from low to high relevance, and (3) reversing the encoded order. We keep the position of the sink block consistent in all experiments.


\begin{table}[ht]
\centering
\small
\setlength{\tabcolsep}{3pt} 
\renewcommand{\arraystretch}{1.1} 
\begin{tabular}{lllllll}
\toprule
& \multicolumn{3}{c}{\textbf{\abbrours}} & \multicolumn{3}{c}{\textbf{Non-cached}} \\
\textbf{Dataset} & \textbf{In.} & \textbf{L-to-H} & \textbf{Rev.} & \textbf{In.} & \textbf{L-to-H} & \textbf{Rev.}\\
\midrule
Banking   & 0.807 & 0.798 & 0.737$^*$ & 0.821 & 0.828 & 0.811 \\
Clinic    & 0.846 & 0.832 & 0.772$^*$ & 0.868 & 0.863 & 0.854 \\
NLU       & 0.832 & 0.838 & 0.804$^*$ & 0.848 & 0.855 & 0.850 \\
TREC      & 0.923 & 0.927 & 0.914 & 0.929 & 0.931 & 0.928 \\
TREC Fine  & 0.759 & 0.782 & 0.738 & 0.845 & 0.850 & 0.840 \\
\midrule
Avg     & 0.834 & 0.835 & 0.794$^*$ & 0.862 & 0.865 & 0.857 \\
\bottomrule
\end{tabular}
\caption{\label{tab:ordering-effect}
Comparison of block ordering strategies with Llama-2 30k context. $^*$ indicates statistical significance (p<0.05) in a T-test against the original encoded order (\textit{In.}). \textit{L-to-H} represents ordering from low to high relevance, and \textit{Rev.} refers to reverse ordering.
}
\end{table}

As shown in Table~\ref{tab:ordering-effect}, reversing the order of cached embeddings significantly decreases performance. We believe this ordering dependence is because each block is permitted a small amount of local attention (i.e. to the previous two blocks) when encoding, imposing a loose ordering of blocks.


\section{Related Work}

\textbf{Many-shot ICL.}
While our experiments focus on classification datasets, many-shot ICL has demonstrated potential across a wide range of tasks, including sequential parity, sentiment analysis, summarization, and translation \cite{agarwal2024many,li2023incontext,bertsch2024LC-ICL}. \citet{agarwal2024many} showed it to be comparable to finetuning in translation, while \citet{yin2024ICL} found it more effective than finetuning in datasets with implicit structures. These work motivate many-shot ICL beyond our considered datasets and models.

\textbf{ICL demonstration selection.} 
ICL demonstration selection is an active area of research. Key approaches include demonstration augmentation, demonstration pool curation, retrieval ICL \cite{dong2024survey}. The effectiveness of retrieval ICL is influenced by multiple factors. However, \citet{luo2024retrievedICL} shows that a simple and efficient BM25 retriever is effective across datasets, performing within 0.5\% of a dual encoder-based retriever (GTR) across five tasks. To improve retrieval efficiency, CPU-based retrieval-augmented generation (RAG) techniques and approximate nearest neighbor (ANN) indexing \cite{malkov2018efficient} can also be incorporated into \abbrours.

\textbf{Sparse Attention Mechanisms} have been explored extensively in Transformers and LLMs to increase efficiency, with various patterns such as block-wise sparse attention \cite{child2019generating,  zaheer2020big, wang2024precision,acharya2024starattention}, hierarchical attention \cite{yang2016hierarchical}, and layer or head-dependent sparsity. These methods primarily aim to improve training and inference efficiency together. However, our approach focus on efficiency without training.

A related training-free approach is Parallel Context Windows \cite{ratner2023parallel}, which also uses block-sparse attention. However, its goal is different: to extend the context length of limited-context models by reusing the same positional embeddings across blocks. This results in a weaker baseline compared to our approach, which operates on models that already support long contexts.

Among existing approaches, \citet{Lu2024TurboRAGAR} and \citet{sun2024blockattentionefficientrag} are most similar to our method, but in the RAG setting. Like our approach, they incorporate a sparse attention encoding phase and optimize key-value (KV) cache re-utilization. However, a key difference is that these methods require fine-tuning to adapt to the sparse attention pattern. As discussed in Section \ref{sec:sparse attention}, we hypothesize that this arises due to the absence of a local context component in their design. 

\textbf{Key-value (KV) cache compression.} A common approach is token-level KV cache eviction to reduce memory overhead and speedup inference \cite{xiao2024efficient, xiao2024duo, Zhang2023H2OHO}. StreamingLLM \cite{xiao2024efficient} discovered the "attention sink" phenomenon and popularized token eviction by retaining only the initial and recent KV pairs. More recent methods \cite{li2024snapkv,Zhang2023H2OHO} refine this approach by selectively retaining some KV pairs. However, eviction-based strategies are not well suited for many-shot ICL, as different test queries benefit from attending to distinct sets of demonstrations.

Beyond eviction, other KV cache compression methods focus on quantization and low-rank approximation to reduce storage requirements while retaining all tokens \cite{liu2024minicachekvcachecompression,zhang2024unifyingkvcachecompression}. These strategies can potentially be integrated with \abbrours to achieve further memory reductions while maintaining inference quality.


\section{Conclusion}

We expect that many-shot ICL will be most useful in settings where the available data may change over time, either because of liability and privacy concerns \cite{liu2024rethinkingmachineunlearninglarge} or because of the continual addition of data in an online learning setting \cite{hoi2018onlinelearningcomprehensivesurvey}. Unlike traditional cached ICL or finetuning, the cost of setup when adding additional demonstrations to the demonstration pool using \ours is linear, and the increased cost at inference time is marginal, depending on the retrieval method used. 

By significantly improving inference efficiency while preserving accuracy, we hope that \abbrours opens the door for adoption of many-shot ICL in real-world applications. 



\section{Limitations}

The exact tradeoff between finetuning, \fixedicl, and \ours depends on the context length and details of the model chosen (e.g. the use of GQA \cite{ainslie2023gqa}); it is possible that there are some combinations of model, context length, and dataset where the marginal performance and efficiency gains of \abbrours may be small or even nonexistent.  
For instance, because our method uses retrieval-based ICL, tasks where retrieval-based ICL is ineffective also will not benefit from our framework. In particular, tasks that require synthesizing information from all demonstrations (e.g. estimating the percentage of Amazon reviews that are positive \cite{shaham-etal-2023-zeroscrolls}) will not benefit from these efficiency speedups.

\section*{Acknowledgments}
AB was supported by a grant from the National Science Foundation Graduate Research Fellowship Program under Grant No. DGE2140739.
Any opinions, findings, and conclusions or recommendations expressed in this material are those of the author(s) and do not necessarily reflect the views of the sponsors.




\clearpage
\appendix
\section{Efficiency metrics}
\label{appendix:efficiency}

In main results \S~\ref{sec:main results} we show the relative speedup. Here we report the raw metrics given our computational enviornment. The inference latency is average of 5 runs and rounded, its standard deviation is within 0.1. We evaluate inference with Huggingface transformers \cite{wolf2020transformers} using Flash Attention 2 \cite{dao2024flashattention2}, or Flex Attention \cite{dong2024flexattention} for \abbrours pre-encoding.

\begin{table}[ht]
\centering
\small
\setlength{\tabcolsep}{6pt} 
\renewcommand{\arraystretch}{1.5}
\resizebox{\columnwidth}{!}{ 
\begin{tabular}{lcc}
\toprule
\textbf{Method} & \textbf{\makecell[c]{Setup Time}} & \textbf{\makecell[c]{Inference Latency \\ (queries/s)}} \\
\midrule
\multicolumn{3}{l}{\textbf{30k Context Length w/ Llama-2-7B w/ single L40S GPU}} \\
\midrule
RetICL (no cache)  & 2 sec  & 0.8  \\
Fixed ICL (cached) & 9 sec  & 1.5  \\
Finetuning       & > 20 min  & 6.6  \\
\abbrours         & 6 sec  & 3.4  \\
\midrule
\multicolumn{3}{l}{\textbf{30k Context Length w/ Llama-3.1-8B w/ single L40S GPU}} \\
\midrule
RetICL (no cache)  & 2 sec  & 1.3  \\
Fixed ICL (cached) & 10 sec  & 11.6  \\
Finetuning       & > 20 min  & 15.8  \\
\abbrours         & 6 sec  & 12.8  \\
\midrule
\multicolumn{3}{l}{\textbf{90k Context Length w/ Llama-3.1-8B w/ single A100 GPU}} \\
\midrule
RetICL (no cache)  & 2 sec  & 0.4  \\
Fixed ICL (cached) & 13 sec  & 6.8  \\
Finetuning       & > 40 min  & 8.9  \\
\abbrours         & 7 sec  & 7.7  \\
\bottomrule
\end{tabular}
}
\caption{Setup time and inference latency (queries/s) across different methods and context lengths on Llama models with specified GPUs.}
\label{tab:app_latency_comparison}
\end{table}

\section{Finetuning Setup}
\label{appendix:finetune}

We follow the same setup as \citet{bertsch2024LC-ICL} for LoRA finetuning, using Hugging Face Transformers \cite{wolf2020transformers} along with the peft package \cite{peft}. All finetuning is run on a single L40S GPU (48G). We finetune for 30 epochs, with LoRa rank $r=8$, scaling factor $\alpha = 32$, and dropout $0.1$. We use a learning rate of $0.001$, a batch size of $32$, and a weight decay of $0.01$.
\section{Accuracy Trend}
\label{appendix:accuracy}

\begin{figure}[h]
    \centering
    \includegraphics[width=0.9\linewidth]{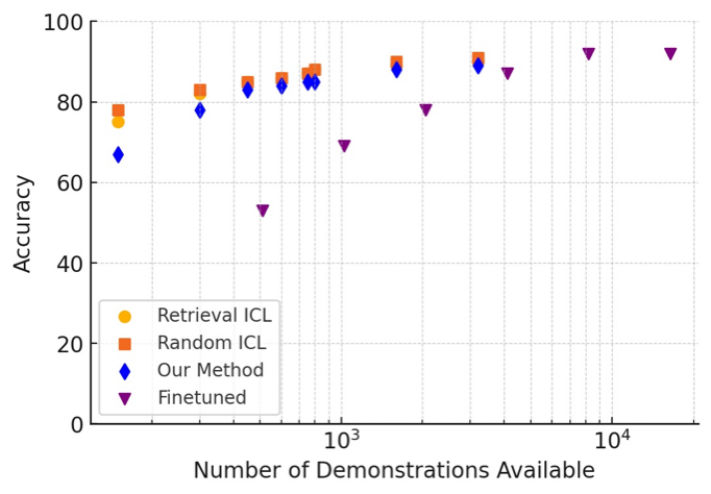}
    \caption{Performance trend for Banking 77.}
    \label{fig:banking77-acc-trend}
    
    \vspace{1em} 
    
    \includegraphics[width=0.9\linewidth]{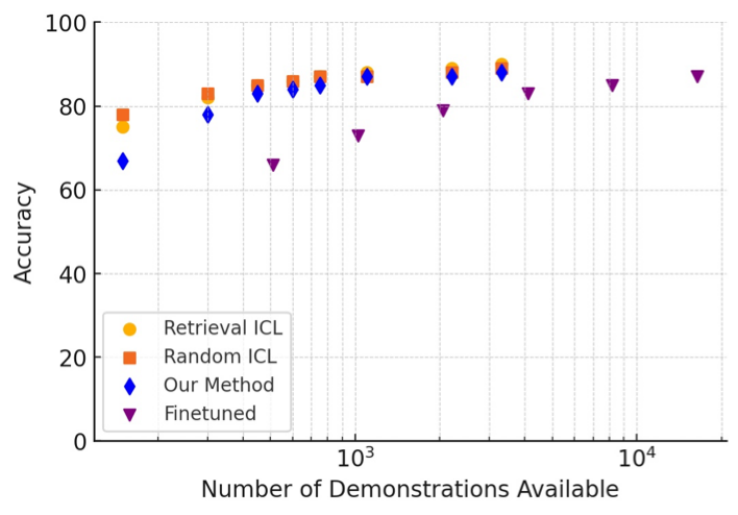}
    \caption{Performance trend for NLU.}
    \label{fig:nlu-acc-trend}
\end{figure}

In the main results (\S~\ref{sec:main results}), we consider only the 30k (about 1k demos) or 90k (about 3k demos) demonstration pool. In Figures~\ref{fig:banking77-acc-trend} and \ref{fig:nlu-acc-trend}, we report two examples of the accuracy trend for each method as the size of demonstration pool increases. The trends we describe at 30k and 90k are consistent at other context lengths, although \abbrours seems slightly less effect at very short context lengths; we do not believe this is a concern, as the relative efficiency gain from sparsity is minimal in short-context ICL.

\end{document}